\documentclass[a4paper]{article}

\usepackage{INTERSPEECH2016}

\usepackage{graphicx}
\usepackage{amssymb,amsmath,bm}
\usepackage{textcomp}
\usepackage[printonlyused,nolist]{acronym}
\usepackage{flushend} 
\usepackage{lipsum}
\usepackage{afterpage}
\usepackage{dblfloatfix}
\usepackage{tikz}
\usepackage{pgfplots}

\ninept

\title{An improved uncertainty decoding scheme with weighted samples for DNN-HMM hybrid systems}

\makeatletter
\def\name#1{\gdef\@name{#1\\}}
\makeatother \name{{\em Christian Huemmer$^1$, Ram\'on Fern\'andez Astudillo$^2$, and Walter Kellermann$^1$}}
  \address{$^1$Multimedia Communications and Signal Processing,\\ University of Erlangen-Nuremberg, Erlangen, Germany \\
  $^2$Spoken Language Systems Laboratory, INESC-ID-Lisboa, Lisboa, Portugal \\
  {\small \tt \{huemmer,wk\}@lnt.de, ramon@astudillo.com}
}

\begin{acronym}
\acro{CDR}{coherent-to-diffuse power ratio}
\acro{GMMs}{Gaussian mixture models}
\acro{DNNs}{deep neural networks}
\acro{ASR}{automatic speech recognition}
\acro{HMM}{hidden Markov model}
\acro{MCE}{minimum classification error}
\acro{DOA}{direction of arrival}
\acro{STFT}{short-time Fourier transform}
\acro{MVDR}{minimum variance distortionless response}
\acro{STFT-UP}{STFT uncertainty propagation}
\acro{PDF}{probability density function}
\acro{MVN}{mean and variance normalization}
\acro{WER}{word error rate}
\acro{GPU}{graphics processing unit}
\acro{MFCCs}{mel-frequency cepstral coefficients}
\end{acronym}

\begin{document}

  \maketitle
  \begin{abstract}
      In this paper, we advance a recently-proposed uncertainty decoding scheme for DNN-HMM (deep neural network - hidden Markov model) hybrid systems.
      This numerical sampling concept averages DNN outputs produced by a finite set of feature samples (drawn from a probabilistic distortion model) to approximate the posterior likelihoods of the context-dependent HMM states.
      As main innovation, we propose a weighted DNN-output averaging based on a minimum classification error criterion and apply it to a
      probabilistic distortion model for spatial diffuseness features.
      The experimental evaluation is performed on the $8$-channel REVERB Challenge task using a DNN-HMM hybrid system with multichannel front-end signal enhancement. We show that the recognition accuracy of the DNN-HMM hybrid system improves by incorporating uncertainty decoding based on random sampling and that the proposed weighted DNN-output averaging further reduces the word error rate scores.
  \end{abstract}
\section{Introduction}
Since \ac{DNNs} have become an important part of state-of-the art \ac{ASR} systems, the mismatch between training and test conditions (e.g., caused by environmental distortions)
motivated various schemes for robust DNN-based speech recognition~\cite{deng_2013,delcroix_2013,Yoshioka_2015}.
For instance, front-end techniques aim at suppressing background noise and reverberation~\cite{martin_2001,cohen_2003,krueger_2010,gales_2011}, while feature transformations reduce irrelevancy and increase discriminability~\cite{Li_2002,li_2005,seide_2011}. Furthermore, back-end techniques adjust acoustic model parameters~\cite{li_2006,liao_2013,yu_2013}.\\
Uncertainty decoding bridges front-end processing and back-end adaptation by combining a probabilistic distortion model with the acoustic model of an \ac{ASR} system~\cite{droppo_2002,kolossa_2005,Liao_2008}. According to this principle, we model acoustic features as random variables and estimate the resulting posterior likelihoods of the context-dependent \ac{HMM} states. However, nonlinear feature transformations in DNN-based \ac{ASR} systems (e.g., sigmoid activation functions) preclude analytical closed-form solutions, which motivated approximate solutions based on numerical sampling techniques~\cite{ahmed_2015,huemmer_2015} and piece-wise function approximations~\cite{ahmed_2015,astudillo_2011}. 
In this paper, we follow uncertainty decoding based on random sampling~\cite{huemmer_2015}, as it has been shown to improve the accuracy of DNN-based \ac{ASR} systems even for a small number of samples~\cite{huemmer_2015,huemmer_2016}. This strategy averages the DNN outputs produced by a finite set of feature samples drawn from a probabilistic distortion model.
As main innovation, we advance uncertainty decoding based on random sampling by introducing a weighted (instead of arithmetic) DNN-output averaging, where the weights are derived from a \ac{MCE} criterion~\cite{juang_1997,matton_2010}.
Furthermore, we apply this numerical sampling concept to a new probabilistic distortion model for spatial diffuseness features derived by considering the variances of the microphone pair-specific diffuseness feature estimates.\\
The experimental part focuses on the $8$-channel REVERB Challenge task and a DNN-HMM hybrid system trained on multi-condition data. 
As input features of the DNN, we append logmelspec features (extracted from a beamformer output signal) with delta coefficients and add spatial diffuseness features. It is shown that the recognition accuracy improves by incorporating uncertainty decoding based on random sampling and that the proposed weighted DNN-output averaging further reduces the \ac{WER} scores, especially for real-world recordings.\\
This paper is structured as follows: We propose a probabilistic distortion model for spatial diffuseness features in Section~\ref{sec:Theory} and the uncertainty decoding scheme with weighted samples in Section~\ref{sec:TheoryUD}. The DNN-HMM hybrid system is introduced in Section~\ref{sec:DNN_HMM},
followed by the experimental results for the $8$-channel REVERB challenge task in Section~\ref{sec:Exper}. Finally, concluding remarks are given in Section~\ref{sec:Concl}.
%
\begin{figure*}[t]
\normalsize
  \begin{equation}
    \mathit{CDR}_{\nu,n} = \frac{\Gamma_{\text{diff},\nu} \;\text{Re}\{\Gamma_{\nu,n}\} -{|\Gamma_{\text{diff},\nu}|}^2 - \sqrt{\Gamma_{\text{diff},\nu}^2\, {\text{Re}\{\Gamma_{\nu,n}\}}^2 - \Gamma_{\text{diff},\nu}^2\, {|\Gamma_{\nu,n}|}^2 + \Gamma_{\text{diff},\nu}^2 - 2\, \Gamma_{\text{diff},\nu}\, \text{Re}\{\Gamma_{\nu,n}\} + {|\Gamma_{\nu,n}|}^2}}{{|\Gamma_{\nu,n}|}^2 - 1}
    \label{eq:CDR_estimator}
  \end{equation}
	\vspace{2mm}
  \hrulefill
\end{figure*}
\section{Probabilistic distortion model for spatial diffuseness features}
\label{sec:Theory}
%
We focus on a multichannel \ac{ASR} system with front-end processing in the \ac{STFT} domain (DFT length $512$). For each microphone pair, indexed by \mbox{$m=1,...,M$}, the real-valued spatial diffuseness \mbox{$0<D^{(m)}_{\nu,n}<1$} in STFT band $\nu=1,...,257$ and at time~$n$ is estimated as described in~\cite{schwarz_sp_2014}:
The \ac{CDR} $\mathit{CDR}^{(m)}_{\nu,n}$ is determined by inserting the estimated complex-valued coherence $\Gamma^{(m)}_{\nu,n}$ (see (6) in~\cite{schwarz_2015}) and the spatial coherence function of a spherically isotropic sound field $\Gamma^{(m)}_{\text{diff},\nu}$ (see (11) in~\cite{schwarz_2015})
into (\ref{eq:CDR_estimator}), where $\text{Re}\{\cdot\}$ is the real part.
Note that the \ac{CDR} estimation in (\ref{eq:CDR_estimator}) is independent of the \ac{DOA}~\cite{schwarz_2015} and that also other CDR estimates could be applied (see overview in~\cite{schwarz_2015}).
The spatial diffuseness is finally obtained by inserting $\mathit{CDR}^{(m)}_{\nu,n}$ into
\begin{equation} 
  D^{(m)}_{\nu,n} = (1 + \mathit{CDR}^{(m)}_{\nu,n})^{-1}.
\label{equ:DiffS}
\end{equation}
From this, the diffuseness feature vector $\mathbf{x}^{(m)}_n$ for each microphone pair is calculated by weighting the length-$257$ vector
\begin{equation}
  \mathbf{d}^{(m)}_n   =  [D^{(m)}_{1,n}, ... , D^{(m)}_{257,n}]^\text{T}
\end{equation}
by the mel-filterbank matrix $\mathbf{W}_\text{mel}$ of dimensions $24 \times 257$:
\begin{equation}
  \mathbf{x}^{(m)}_n   =  [x^{(m)}_{1,n}, ... , x^{(m)}_{24,n}]^\text{T} =  \mathbf{W}_\text{mel} \; \mathbf{d}^{(m)}_n.
\end{equation}
These microphone-pair specific estimates are averaged to determine the spatial diffuseness feature vector $\mathbf{x}_n$:
\begin{equation}
\mathbf{x}_n  = \frac{1}{M}\sum\limits_{m=1}^{M}  \mathbf{x}^{(m)}_n.
\label{equ:DiffAv}
\end{equation}
%
As measure for the uncertainty of the spatial diffuseness feature vector in~(\ref{equ:DiffAv}),
we introduce the Gaussian distribution
\begin{equation}
p(\mathbf{z}_{n}|\mathbf{x}_{n}) = \mathcal{N}\{\mathbf{x}_{n}, \mathbf{V}_{n}\}
\label{equ:PDFz}
\end{equation}
with mean vector $\mathbf{x}_{n}$ and covariance variance
\begin{align}
&\mathbf{V}_n = \text{diag} \{v_{1,n}, ... , v_{24,n}),\\
&v_{\kappa,n} = \frac{1}{M-1}\sum\limits_{m=1}^{M}  \left(x^{(m)}_{\kappa,n}-x_{\kappa,n} \right)^2,
\label{equ:Diffvar}
\end{align}
where $\text{diag}\{\cdot\}$ creates a diagonal matrix from the scalar variances $v_{\kappa,n}$ ($\kappa = 1,...,24$).
Note that we choose a diagonal covariance matrix $\mathbf{V}_{n}$ in (\ref{equ:PDFz}) for a computationally efficient realization of the uncertainty decoding scheme described in the following section.\\
%
\section{Improved uncertainty decoding with weighted samples}
\label{sec:TheoryUD}
\begin{figure}[!b]
\includegraphics[width=0.47\textwidth]{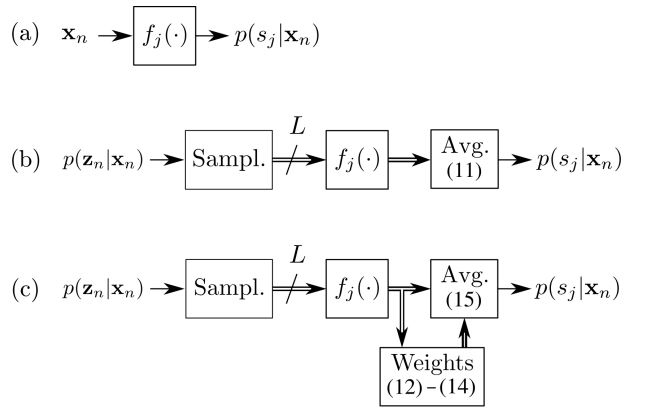}
\caption{Calculation of the posterior likelihood $p(s_j |\mathbf{x}_n )$ in a DNN-HMM hybrid system (a) without and with uncertainty decoding based on random sampling using (b)~arithmetic and (c)~weighted DNN-output averaging.}
\label{fig:Sampl}
\end{figure}

As illustrated in Fig.~\ref{fig:Sampl}(a), the posterior likelihood of the $j$th context-dependent HMM state $s_j$ in the decoding process of a DNN-HMM hybrid system is given by a nonlinear transformation of the feature vector $\mathbf{x}_n$ at time instant~$n$:
\begin{equation}
p(s_j|\mathbf{x}_n) = f_j(\mathbf{x}_n).
\label{equ:Post1}
\end{equation}
Assuming distortions of the feature vector $\mathbf{x}_n$ (reflecting, e.g., measurement uncertainty) to be modeled by a latent random variable $\mathbf{z}_n$ and the nonlinear mapping $f_j(\cdot)$ to be known, the posterior distribution in (\ref{equ:Post1}) reads according to~\cite{huemmer_2015}:
\begin{align}
  p(s_j|\mathbf{x}_n) =\mathcal{E}\{ f_j(\mathbf{z}_n)|\mathbf{x}_n \}.
\label{equ:Post2}
\end{align}
In general, nonlinearities in $f_j(\cdot)$ preclude closed-form solutions of (\ref{equ:Post2}), which motivated
piece-wise function approximations and numerical sampling schemes.
In this paper, we focus on uncertainty decoding based on random sampling, as it has been shown to be promising for improving the accuracy of DNN-based \ac{ASR} systems even for a small number of samples~\cite{huemmer_2015,huemmer_2016}.
As illustrated in Fig.~\ref{fig:Sampl}(b), we draw $L$ samples $\mathbf{z}^{(l)}_n$ from the estimated \ac{PDF}~$p(\mathbf{z}_n|\mathbf{x}_n)$ and average the resulting DNN outputs:
\begin{align}
p(s_j|\mathbf{x}_n) \approx \frac{1}{L} \sum\limits_{l=1}^L f_j(\mathbf{z}^{(l)}_n),
\label{equ:UDaver}
\end{align}   
where $l=1,..,L$. This numerical sampling scheme is modified in the following
by employing a \ac{MCE} criterion as measure for the reliability of DNN output $f_j(\mathbf{z}^{(l)}_n)$~\cite{juang_1997,matton_2010}:
for each sample $\mathbf{z}^{(l)}_n$, the $g$th (sample index $l$ omitted for simplicity) DNN output is identified as the most probable
 \begin{align}
   g = \underset{j}{\operatorname{argmax}} \;  f_j(\mathbf{z}^{(l)}_n)
 \end{align}  
to determine the misclassification measure $e^{(l)}_n$ as the difference between the most probable and the best competing class~\cite{juang_1997}:
 \begin{align} 
   e^{(l)}_n =  f_{g}(\mathbf{z}^{(l)}_n) - \underset{j\neq g}{\operatorname{max}} \;  f_j(\mathbf{z}^{(l)}_n).
   \label{equ:error}
 \end{align}
In other words, $e^{(l)}_n$ is the sample-specific difference between the posterior likelihoods of the two most probable HMM states and thus a confidence measure for the reliability of the classification. From (\ref{equ:error}), we propose to calculate weights
\\
\begin{figure*}[t]
\includegraphics[width=0.99\textwidth]{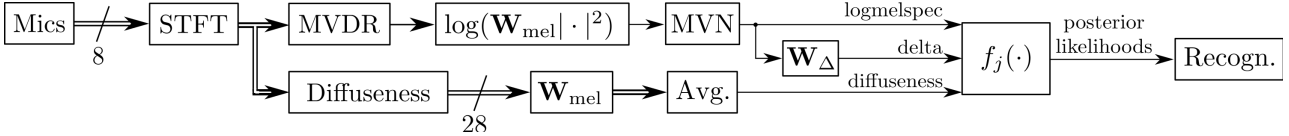}\vspace{-1mm}
\caption{DNN-HMM hybrid system with ``logmelspec+delta+diffuseness'' feature vector being nonlinearly transformed by $f_j(\cdot)$ to the posterior likelihoods of the context-dependent HMM states.
$\mathbf{W}_\text{mel}$ is the mel-filterbank matrix, $\log(\cdot)$ is the natural logarithm, MVN denotes mean and variance normalization and matrix $\mathbf{W}_\Delta$ creates the delta coefficients. For extracting the diffuseness features, we consider $M=28$ non-redundant microphone pairs of the $8$-channel circular microphone array provided by the REVERB Challenge.}
\label{fig:DNN_HMM}
\end{figure*}
%
\begin{table*}[b]
  \caption{\label{tab:Clean} {\it WER scores for the REVERB challenge evaluation test set using different feature types.}}\vspace*{1mm}
   \centerline{
   \begin{tabular}{|c|c|c|c|c|c|c|c|c|c|c|c|c|}
    \hline
   & \multicolumn{7}{|c|}{SimData} & \multicolumn{3}{|c|}{RealData} \\ 
    \hline
   & \multicolumn{2}{|c|}{$T_{60}\approx 0.25$~s} & \multicolumn{2}{|c|}{$T_{60}\approx 0.5$~s} &\multicolumn{2}{|c|}{$T_{60}\approx 0.75$~s}& & \multicolumn{2}{|c|}{$T_{60}\approx 0.7$~s} & \\ 
    \hline
						     & Near        &  Far         & Near         & Far          & Near  	  & Far          & Avg.	        & Near         & Far         & Avg. \\ 
            \hline
            Logmelspec+delta+acceleration(1 mic)     & 5.7         & 6.5          & 6.9          & 11.2         & 8.1 	          & 13.0 	 & 8.6          & 21.2         & 23.0        & 22.1 \\ 
           \hline
           Logmelspec+delta+acceleration(MVDR)       & {\bf 4.7}   & 5.2          & {\bf4.9}     & 6.9          & 6.1    	  & 8.6 	 & 6.1 	        & 13.6         & 17.2        & 15.4 \\ 
           \hline
            Logmelspec+delta(MVDR)+diffuseness       & 4.9         & {\bf 5.1}    & 5.0          & {\bf 6.2}    & {\bf 6.0}       & {\bf 8.0}    & {\bf 5.9}	& {\bf 13.1}   & {\bf 15.3}  & {\bf 14.2} \\
           \hline 
          \end{tabular}
       }
\end{table*}
\begin{align} 
   \omega^{(l)}_n =   e_n^{(l)} / \sum_{l=1}^L e_n^{(l)}
 \end{align}
used for a weighted DNN-output averaging: 
\begin{equation}
  p(s_j |\mathbf{x}_n ) \approx \sum\limits_{l=1}^L  \omega_n^{(l)} f_j(\mathbf{z}^{(l)}_n).
\label{equ:UDaverW}
\end{equation}
An overview of the proposed uncertainty decoding scheme with weighted samples is illustrated in Fig.~\ref{fig:Sampl}(c).\\
Finally, the relation between uncertainty decoding with weighted samples and multistream fusion should be clarified. The latter is (motivated by the human behavior dealing with unexpected data) based on experimental evidence that the error probability in human decoding is given by the product of error probabilities in different frequency bands~\cite{hermansky_2013}. Although this is conceptually different from drawing samples from a \ac{PDF}, the strategies for fusing multiple streams~(e.g., \cite{mesgarani_2011,variani_2013}) might also be of interest for improving the DNN-output averaging as part of uncertainty decoding with weighted samples.
%
\section{DNN-HMM hybrid system}
\label{sec:DNN_HMM}
Besides the spatial diffuseness features of Section~\ref{sec:Theory}, we extract logmelspec features and corresponding delta coefficients which have been frequently used in state-of-the art DNN-HMM hybrid systems  (often termed ``log-filterbank'' features).
As shown in Fig.~\ref{fig:DNN_HMM}, the STFT-domain microphone signals are processed by an MVDR beamformer, transformed into the lower-dimensional logmelspec domain ($24$ coefficients) and normalized by applying per-utterance \ac{MVN}.
We append the delta coefficients and the spatial diffuseness features to create the ``logmelspec+delta+diffuseness'' feature vector of length $72$, which is passed through the nonlinearity $f_j(\cdot)$ realized in our implementation as follows:
	   \begin{itemize}
	    \item Context extension using $\pm 5$ frame splicing (the size of the context window has been manually optimized),
	    \item DNN: $6$ hidden layers, each with $2048$ sigmoid activation functions, output layer with softmax nonlinearity and $3463$ elements (context-dependent HMM states).
	    \end{itemize}
It should be emphasized that Fig.~\ref{fig:DNN_HMM} provides an overview of the DNN-HMM hybrid system without reflecting the probabilistic distortion model of (\ref{equ:PDFz}). As the MVDR beamformer is designed to let plane waves coming from the desired look direction pass the system undistorted~\cite{VanTrees_2004}, we model logmelspec features and respective delta coefficients as deterministic point estimates without observation uncertainty. Thus, the ``logmelspec+delta+diffuseness'' feature vector samples used for uncertainty decoding consist of
$L$ diffuseness feature vector samples drawn from (\ref{equ:PDFz}) appended by the logmelspec features and respective delta coefficients (equal for all samples).
 \begin{table*}[t]
  \caption{\label{tab:Sampl} {\it WER scores for the REVERB challenge evaluation test set achieved by the DNN-HMM hybrid system without (``No UD'') and with uncertainty decoding based on random sampling ($L=30$ samples) using arithmetic~(``UD arithm'') and weighted~(``UD weight'') DNN-output averaging.}}\vspace{1mm}
   \centerline{
   \begin{tabular}{|c|c|c|c|c|c|c|c|c|c|c|c|c|}
    \hline
   & \multicolumn{7}{|c|}{SimData} & \multicolumn{3}{|c|}{RealData} \\ 
    \hline
   & \multicolumn{2}{|c|}{$T_{60}\approx 0.25$~s} & \multicolumn{2}{|c|}{$T_{60}\approx 0.5$~s} &\multicolumn{2}{|c|}{$T_{60}\approx 0.75$~s}& & \multicolumn{2}{|c|}{$T_{60}\approx 0.7$~s} & \\ 
    \hline
                     & Near      &  Far    & Near      & Far    & Near   & Far       & Avg.	    & Near       & Far         & Avg. \\ 
            \hline
No UD                & {\bf 4.9} &  5.1    & 5.0       & 6.2    & 6.0    &  8.0      & 5.9      &  13.1      &  15.3       &  14.2 \\
           \hline
UD arithm ($L=30$)   & 5.0       & 5.0 & {\bf 4.9}       & 5.9    & 5.9    &  7.7      & 5.7    	& 12.9       & 14.9        & 13.9 \\
           \hline
UD weight ($L=30$) & {\bf 4.9} & {\bf 4.9} & {\bf 4.9} & {\bf 5.7} & {\bf 5.8} & {\bf 7.5} &  {\bf 5.6}	& {\bf 12.7}   & {\bf 14.6}  & {\bf 13.6} \\
           \hline 
          \end{tabular}
       }
\end{table*}
%
\section{Experiments}
\label{sec:Exper}
We choose the Kaldi Toolkit~\cite{povey_2011} as ASR back-end system to evaluate the DNN-HMM hybrid system on the $8$-channel REVERB Challenge task~\cite{kinoshita_2013} (WSJ0 trigram 5k language model, circular microphone array with a microphone spacing of $8$~cm).
As first step, a GMM-HMM system is trained on the clean \mbox{WSJCAM0} Cambridge Read News REVERB corpus~\cite{Robinson_2013} with feature extraction following the \mbox{Type-I} creation in~\cite{Rath_2013}, which is state-of-the art in the Kaldi recipe~\cite{povey_2011}. Then, we create a state-frame alignment to train the DNN
on the multi-condition training sets (each of $7861$ utterances) provided by the \mbox{REVERB} challenge~\cite{kinoshita_2013}. This is realized using ``Karel's implementation`` of the Kaldi Toolkit including generative pretraining (on restricted Bolzmann machines) and discriminative fine-tuning (using a mini-batch stochastic gradient descent approach)~\cite{hinton_2012}. 
It should be emphasized that the front-end enhancement and feature extraction is identical
during training and testing.\\
\noindent The evaluation test set of the REVERB-Challenge task ($\sim5000$ environmentally-distorted utterances) consists of ...
\begin{itemize}
 \item artificially corrupted data (``SimData'') using measured impulse responses (\mbox{$T_{60}\approx 0.25$~s, $0.5$~s and $0.7$~s}), recorded noise sequences (added to the microphones signals with a signal-to-noise ratio of $20$~dB) and source-microphone spacings of $0.5$~m (``Near'') and $2$~m (``Far''),
 \item multichannel recordings (``RealData'') in a reverberant ($T_{60}\approx 0.7$~s) and noisy environment with source-microphone spacings of $1$~m (``Near'') and $2.5$~m (``Far'').
\end{itemize}
As first experiment, we evaluate the performance of the DNN-HMM hybrid system without uncertainty decoding. Here we compare the recognition accuracy achieved by ``Logmelspec+delta+diffuseness'' features (see Fig.~\ref{fig:DNN_HMM}) to the logmelspec features with respective delta and acceleration coefficients extracted from a single microphone signal \mbox{``Logmelspec+delta+acceleration(1 mic)''} and the beamformer output signal \mbox{``Logmelspec+delta+acceleration(MVDR)''}.
It is obvious from Table~\ref{tab:Clean}, that spatial filtering and replacing acceleration coefficients by diffuseness features significantly improves the recognition accuracy of the DNN-HMM hybrid system, especially in scenarios with large source-microphone spacing and for real-world recordings.\\
\begin{figure}[!b]
\centering
\begin{tikzpicture}[scale=1]
\def\lx{-0.3}
\def\ly{3.2}
\begin{axis}[
      width=7cm,height=4.5cm,grid=major,grid style = {dotted,black},
      ylabel={$\text{WER} \;/$ \;\% $\; \rightarrow$},
      xlabel={Number of $L$ samples $\; \rightarrow$},
      ymin=13, ymax=15.5,xmin=1,xmax=60,
      ]
      \addplot[thick,green,solid] table [x index=0, y index=1]{UDnew.dat};
      \addplot[thick,black,dashed] table [x index=0, y index=1]{UDold.dat};
      \addplot[thick,red,solid] table [x index=0, y index=1]{NoUD.dat};
 \end{axis}
 \draw[fill=white] (\lx,\ly) rectangle (6.0+\lx,0.62+\ly);
 \draw[red,thick,solid] (0.2+\lx,0.25+\ly) -- +(0.3,0) node[anchor=mid west,black] {\small No UD};
 \draw[green,thick,solid] (4.0+\lx,0.25+\ly) -- +(0.3,0) node[anchor=mid west,black] {\small UD weight};
 \draw[black,thick,dashed] (1.95+\lx,0.25+\ly) -- +(0.3,0) node[anchor=mid west,black] {\small UD arithm};
 \end{tikzpicture}
\caption{WER scores for the real-world recordings (averaging the scenarios ``Near'' and ``Far'') of the REVERB challenge evaluation test set achieved by the DNN-HMM hybrid system without (``No UD'') and with uncertainty decoding based on random sampling using arithmetic~(``UD arithm'') and weighted (``UD weight'') DNN-output averaging.}
 \label{fig:ResL}
 \end{figure}
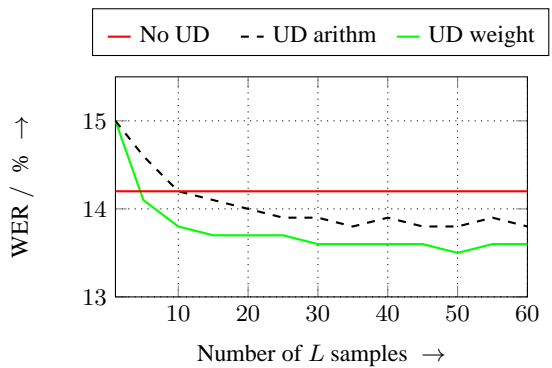
\hspace*{-2mm} Next, we compare the recognition accuracy achieved by DNN-HMM hybrid system using uncertainty decoding
based on~...
\begin{itemize}
\item the arithmetic DNN-output averaging in (\ref{equ:UDaver}), termed \mbox{``UD arithm''},
\item the weighted DNN-output averaging in (\ref{equ:UDaverW}), denoted as ``UD weight''.
\end{itemize}
As in other approaches deriving uncertainty from front-end enhancement~\cite{delcroix2008combined,kolossa2010independent},  we employ a scaling factor in~(\ref{equ:Diffvar}) (manually optimized to a value of $0.1$) to compensate for the inaccuracies of the variance estimation. \\
First, the impact of the number of $L$ samples on the recognition accuracy of the DNN-HMM hybrid system is evaluated with focus on the practically relevant case of real-world recordings.
As shown in Fig.~\ref{fig:ResL}, the \ac{WER} scores for real data (averaging the scenarios ``Near'' and ``Far'') decrease with increasing value of $L$, while already a small number of samples is sufficient for improving the recognition accuracy compared to the DNN-HMM hybrid system without uncertainty decoding (``No UD'').
Further, replacing arithmetic (``UD arithm'' in Fig.~\ref{fig:ResL}) by weighted (``UD weight'' in Fig.~\ref{fig:ResL})  DNN-output averaging leads to a consistent reduction of the WER scores.
It should be mentioned that no significant further improvement was observed by increasing the number of samples above $L=60$.
In summary, the results in Table~\ref{tab:Sampl} for $L=30$ samples emphasize the performance gain achieved by applying uncertainty decoding based on random sampling using the proposed weighted DNN-output averaging.\\
It is worth highlighting that uncertainty decoding based on random sampling modifies the posterior-likelihood calculation and leaves the remaining parts of the decoding procedure (e.g., grammar and language model) untouched. As a consequence, the average decoding time in our implementation is only increased by $76\%$ for $L=30$ samples (Intel i7 with $3.4$~GHz, GPU: NVIDIA GeForce GTX 650 Ti) over regular decoding (no uncertainty decoding).\\
%
  \section{Conclusions}
    \label{sec:Concl}
We advance a recently-proposed uncertainty decoding scheme for DNN-HMM hybrid systems which averages posterior likelihoods of context-dependent HMM states produced by a finite set of feature samples.
As main innovation, we introduce a weighted (instead of arithmetic) posterior-likelihood averaging based on a minimum classification error criterion and apply it to a new probabilistic distortion model for spatial diffuseness features.
The experimental results for the $8$-channel REVERB challenge task show that incorporating uncertainty decoding improves the recognition accuracy of a DNN-HMM hybrid with multichannel front-end signal enhancement and that the proposed weighted DNN-output averaging further reduces the word error rate scores.\\

  \section{Acknowledgements}
  
  The authors would like to thank the Deutsche Forschungsgemeinschaft (contract number KE 890/4-2) and the
  Foundation for Science and Technology (project UID/CEC/50021/2013 and grant SFRH/BPD/68428/2010)
	for supporting this work.
	\vspace{0.4cm}
\atColsBreak{\vskip5pt} 
  \eightpt
  \bibliographystyle{IEEEtran}
  \bibliography{mybib}

\begin{thebibliography}{10}
\providecommand{\url}[1]{#1}
\csname url@samestyle\endcsname
\providecommand{\newblock}{\relax}
\providecommand{\bibinfo}[2]{#2}
\providecommand{\BIBentrySTDinterwordspacing}{\spaceskip=0pt\relax}
\providecommand{\BIBentryALTinterwordstretchfactor}{4}
\providecommand{\BIBentryALTinterwordspacing}{\spaceskip=\fontdimen2\font plus
\BIBentryALTinterwordstretchfactor\fontdimen3\font minus
  \fontdimen4\font\relax}
\providecommand{\BIBforeignlanguage}[2]{{%
\expandafter\ifx\csname l@#1\endcsname\relax
\typeout{** WARNING: IEEEtran.bst: No hyphenation pattern has been}%
\typeout{** loaded for the language `#1'. Using the pattern for}%
\typeout{** the default language instead.}%
\else
\language=\csname l@#1\endcsname
\fi
#2}}
\providecommand{\BIBdecl}{\relax}
\BIBdecl

\bibitem{deng_2013}
L.~Deng, J.~Li, J.-T. Huang, K.~Yao, D.~Yu, F.~Seide, M.~Seltzer, G.~Zweig,
  X.~He, J.~Williams, and {others}, ``Recent advances in deep learning for
  speech research at {Microsoft},'' in \emph{Proc. IEEE Int. Conf. Acoustics,
  Speech, Signal Process. (ICASSP)}.\hskip 1em plus 0.5em minus 0.4em\relax
  IEEE, 2013, pp. \mbox{8604--8608}.

\bibitem{delcroix_2013}
M.~Delcroix, Y.~Kubo, T.~Nakatani, and A.~Nakamura, ``Is speech enhancement
  pre-processing still relevant when using deep neural networks for acoustic
  modeling?'' in \emph{Proc. {Interspeech}}, 2013, pp. \mbox{2992--2996}.

\bibitem{Yoshioka_2015}
T.~Yoshioka and M.~Gales, ``Environmentally robust {ASR} front-end for deep
  neural network acoustic models,'' \emph{Computer Speech Language}, vol.~31,
  no.~1, pp. \mbox{65--86}, 2015.

\bibitem{martin_2001}
R.~Martin, ``Noise power spectral density estimation based on optimal smoothing
  and minimum statistics,'' \emph{IEEE Trans. Speech, Audio Process.}, vol.~9,
  no.~5, pp. 504--512, 2001.

\bibitem{cohen_2003}
I.~Cohen, ``Noise spectrum estimation in adverse environments: improved minima
  controlled recursive averaging,'' \emph{IEEE Trans. Speech, Audio Process.},
  vol.~11, no.~5, pp. \mbox{466--475}, Sep. 2003.

\bibitem{krueger_2010}
A.~Krueger and R.~Haeb-Umbach, ``Model-based feature enhancement for
  reverberant speech recognition,'' \emph{IEEE Trans. Audio, Speech, Lang.
  Process.}, vol.~18, no.~7, pp. \mbox{1692--1707}, 2010.

\bibitem{gales_2011}
M.~J.~F. Gales and Y.-Q. Wang, ``Model-based approaches to handling additive
  noise in reverberant environments,'' in \emph{Proc. Joint Workshop Hands-free
  Speech Comm. Microphone Arrays ({HSCMA})}.\hskip 1em plus 0.5em minus
  0.4em\relax IEEE, 2011, pp. \mbox{121--126}.

\bibitem{Li_2002}
B.~Li and K.~C. Sim, ``Comparison of discriminative input and output
  transformations for speaker adaptation in the hybrid {NN/HMM} systems,'' in
  \emph{Proc. {Interspeech}}, 2010, pp. \mbox{526--529}.

\bibitem{li_2005}
L.~Deng, J.~Wu, J.~Droppo, and A.~Acero, ``Analysis and comparison of two
  speech feature extraction/compensation algorithms,'' \emph{IEEE Signal
  Processing Letters}, vol.~12, no.~6, pp. \mbox{477--480}, 2005.

\bibitem{seide_2011}
F.~Seide, G.~Li, X.~Chen, and D.~Yu, ``Feature engineering in context-dependent
  deep neural networks for conversational speech transcription,'' in
  \emph{Proc. {IEEE} {Workshop} Automatic {Speech} {Recognition},
  {Understanding} ({ASRU})}.\hskip 1em plus 0.5em minus 0.4em\relax IEEE, 2011,
  pp. \mbox{24--29}.

\bibitem{li_2006}
X.~Li and J.~Bilmes, ``Regularized adaptation of discriminative classifiers,''
  in \emph{Proc. IEEE Int. Conf. Acoustics, Speech, Signal Process.
  (ICASSP)}.\hskip 1em plus 0.5em minus 0.4em\relax IEEE, 2006, pp.
  \mbox{237--240}.

\bibitem{liao_2013}
H.~Liao, ``Speaker adaptation of context dependent deep neural networks,'' in
  \emph{Proc. IEEE Int. Conf. Acoustics, Speech, Signal Process.
  (ICASSP)}.\hskip 1em plus 0.5em minus 0.4em\relax IEEE, 2013, pp.
  \mbox{7947--7951}.

\bibitem{yu_2013}
D.~Yu, K.~Yao, H.~Su, G.~Li, and F.~Seide, ``{KL}-divergence regularized deep
  neural network adaptation for improved large vocabulary speech recognition,''
  in \emph{Proc. IEEE Int. Conf. Acoustics, Speech, Signal Process.
  (ICASSP)}.\hskip 1em plus 0.5em minus 0.4em\relax IEEE, 2013, pp.
  \mbox{7893--7897}.

\bibitem{droppo_2002}
J.~Droppo, A.~Acero, and L.~Deng, ``Uncertainty decoding with {SPLICE} for
  noise robust speech recognition,'' in \emph{Proc. IEEE Int. Conf. Acoustics,
  Speech, Signal Process. (ICASSP)}.\hskip 1em plus 0.5em minus 0.4em\relax
  IEEE, 2002, pp. \mbox{57--60}.

\bibitem{kolossa_2005}
D.~Kolossa, A.~Klimas, and R.~Orglmeister, ``Separation and robust recognition
  of noisy, convolutive speech mixtures using time-frequency masking and
  missing data techniques,'' in \emph{Proc. IEEE Workshop Appl. Signal Process.
  Audio Acoustics (WASPAA)}.\hskip 1em plus 0.5em minus 0.4em\relax IEEE, 2005,
  pp. \mbox{82--85}.

\bibitem{Liao_2008}
H.~Liao and M.~Gales, ``Issues with uncertainty decoding for noise robust
  automatic speech recognition,'' \emph{Speech Communication}, vol.~50, no.~4,
  pp. \mbox{265--277}, 2008.

\bibitem{ahmed_2015}
A.~Abdelaziz, S.~Watanabe, J.~Hershey, E.~Vincent, and D.~Kolossa,
  ``Uncertainty propagation through deep neural networks,'' in \emph{Proc.
  {Interspeech}}, Sep. 2015, pp. \mbox{3561--3565}.

\bibitem{huemmer_2015}
C.~Huemmer, R.~Maas, A.~Schwarz, R.~Astudillo, and W.~Kellermann, ``Uncertainty
  decoding for {DNN-HMM} hybrid systems based on numerical sampling,'' in
  \emph{Proc. {Interspeech}}, Sep. 2015, pp. 3556--3560.

\bibitem{astudillo_2011}
R.~Astudillo and J.~Neto, ``Propagation of uncertainty through multilayer
  perceptrons for robust automatic speech recognition,'' in \emph{Proc.
  Interspeech}, 2011, pp. \mbox{461--464}.

\bibitem{huemmer_2016}
C.~Huemmer, A.~Schwarz, R.~Maas, H.~Barfuß, R.~Astudillo, and W.~Kellermann,
  ``A new uncertainty decoding scheme for {DNN-HMM} hybrid systems with
  multichannel speech enhancement,'' in \emph{Proc. IEEE Int. Conf. Acoustics,
  Speech, Signal Process. (ICASSP)}, Mar. 2016, pp. 5760--5764.

\bibitem{juang_1997}
B.-H. Juang, W.~Hou, and C.-H. Lee, ``Minimum classification error rate methods
  for speech recognition,'' \emph{IEEE Trans. Speech, Audio Process.}, vol.~5,
  no.~3, pp. 257--265, 1997.

\bibitem{matton_2010}
M.~Matton, D.~Van~Compernolle, and R.~Cools, ``Minimum classification error
  training in example based speech and pattern recognition using sparse weight
  matrices,'' \emph{Journal Computational, Applied Mathematics}, vol. 234,
  no.~4, pp. 1303--1311, 2010.

\bibitem{schwarz_sp_2014}
A.~Schwarz, C.~Huemmer, R.~Maas, and W.~Kellermann, ``Spatial diffuseness
  features for {DNN}-based speech recognition in noisy and reverberant
  environments,'' in \emph{Proc. IEEE Int. Conf. Acoustics, Speech, Signal
  Process. (ICASSP)}.\hskip 1em plus 0.5em minus 0.4em\relax IEEE, 2014, pp.
  \mbox{4380--4384}.

\bibitem{schwarz_2015}
A.~Schwarz and W.~Kellermann, ``Coherent-to-diffuse power ratio estimation for
  dereverberation,'' \emph{{IEEE}/{ACM} Trans. Audio, Speech, Language
  Process.}, vol.~23, no.~6, pp. \mbox{1006--1018}, 2015.

\bibitem{hermansky_2013}
H.~Hermansky, ``Multistream recognition of speech: Dealing with unknown
  unknowns,'' \emph{Proceedings IEEE}, vol. 101, no.~5, pp. 1076--1088, 2013.

\bibitem{mesgarani_2011}
N.~Mesgarani, S.~Thomas, and H.~Hermansky, ``Adaptive stream fusion in
  multistream recognition of speech.'' in \emph{Proc. {Interspeech}}, 2011, pp.
  2329--2332.

\bibitem{variani_2013}
E.~Variani, F.~Li, and H.~Hermansky, ``Multi-stream recognition of noisy speech
  with performance monitoring.'' in \emph{Proc. {Interspeech}}, 2013, pp.
  2978--2981.

\bibitem{VanTrees_2004}
H.~Van~Trees, \emph{Detection, Estimation, and Modulation Theory, Part IV,
  Optimum Array Processing}.\hskip 1em plus 0.5em minus 0.4em\relax New York:
  Wiley, 2004.

\bibitem{povey_2011}
D.~Povey, A.~Ghoshal, G.~Boulianne, L.~Burget, O.~Glembek, N.~Goel,
  M.~Hannemann, P.~Motlicek, Y.~Qian, P.~Schwarz, J.~Silovsky, G.~Stemmer, and
  K.~Vesely, ``The {Kaldi} speech recognition toolkit,'' in \emph{IEEE 2011
  Workshop Automatic Speech Recognition, Understanding}.\hskip 1em plus 0.5em
  minus 0.4em\relax IEEE, 2011.

\bibitem{kinoshita_2013}
K.~Kinoshita, M.~Delcroix, T.~Yoshioka, T.~Nakatani, A.~Sehr, W.~Kellermann,
  and R.~Maas, ``The {REVERB} challenge: {A} common evaluation framework for
  dereverberation and recognition of reverberant speech,'' in \emph{Proc. IEEE
  Workshop Appl. Signal Process. Audio Acoustics (WASPAA)}.\hskip 1em plus
  0.5em minus 0.4em\relax IEEE, 2013, pp. \mbox{1--4}.

\bibitem{Robinson_2013}
T.~Robinson, J.~Fransen, D.~Pye, J.~Foote, S.~Renals, P.~Woodland, and
  S.~Young, ``{WSJCAM0} {C}ambridge read news for {REVERB} {LDC2013E109},'' Web
  Download, 2013.

\bibitem{Rath_2013}
S.~Rath, D.~Povey, K.~Veseley, and J.~Cernocky, ``Improved feature processing
  for deep neural networks,'' in \emph{Proc. {Interspeech}}, 2013, pp.
  \mbox{109--113}.

\bibitem{hinton_2012}
G.~Hinton, L.~Deng, D.~Yu, A.-R. Mohamed, N.~Jaitly, A.~Senior, V.~Vanhoucke,
  P.~Nguyen, T.~S.~G. Dahl, and B.~Kingsbury, ``Deep neural networks for
  acoustic modeling in speech recognition,'' \emph{IEEE Signal Process.
  Magazine}, vol.~29, no.~6, pp. 82--97, 2012.

\bibitem{delcroix2008combined}
M.~Delcroix, T.~Nakatani, and S.~Watanabe, ``Combined static and dynamic
  variance adaptation for efficient interconnection of speech enhancement
  pre-processor with speech recognizer,'' in \emph{Proc. IEEE Int. Conf.
  Acoustics, Speech, Signal Process. (ICASSP)}.\hskip 1em plus 0.5em minus
  0.4em\relax IEEE, 2008, pp. 4073--4076.

\bibitem{kolossa2010independent}
D.~Kolossa, R.~F. Astudillo, E.~Hoffmann, and R.~Orglmeister, ``Independent
  component analysis and time-frequency masking for speech recognition in
  multitalker conditions,'' \emph{EURASIP Journal Audio, Speech, Music
  Process.}, vol. 2010, pp. 1--13, 2010.

\end{thebibliography}

\end{document}